\documentclass[acmsmall]{acmart}
\usepackage{subfigure}
\usepackage{soul}
\usepackage{color}
\def\BibTeX{{\rm B\kern-.05em{\sc i\kern-.025em b}\kern-.08emT\kern-.1667em\lower.7ex\hbox{E}\kern-.125emX}}
    
\begin{document}

% The "title" command has an optional parameter, allowing the author to define a "short title" to be used in page headers.
\title{\textit{Being the center of attention}: A Person-Context CNN framework for Personality Recognition}

%
% The "author" command and its associated commands are used to define the authors and their affiliations.
% Of note is the shared affiliation of the first two authors, and the "authornote" and "authornotemark" commands
% used to denote shared contribution to the research.
\author{Dario Dotti}
\email{dario.dotti@maastrichtuniversity.nl}
\orcid{}
\affiliation{%
  \institution{Maastricht University}
  \streetaddress{Bouillonstraat 8-10}
  \city{Maastricht}
  \country{The Netherlands}
  \postcode{6211 LH}
}

\author{Mirela Popa}
\email{mirela.popa@maastrichtuniversity.nl}
\affiliation{%
 \institution{Maastricht University}
 \streetaddress{ Bouillonstraat 8-10}
 \city{Maastricht}
 \country{The Netherlands}}
 \postcode{6211 LH}
 
\author{Stylianos Asteriadis}
\email{stelios.asteriadis@maastrichtuniversity.nl}
\affiliation{%
 \institution{Maastricht University}
 \streetaddress{Bouillonstraat 8-10}
 \city{Maastricht}
 \country{The Netherlands}
 \postcode{6211 LH}}

%
% By default, the full list of authors will be used in the page headers. Often, this list is too long, and will overlap
% other information printed in the page headers. This command allows the author to define a more concise list
% of authors' names for this purpose.
\renewcommand{\shortauthors}{Dotti, et al.}

%
% The abstract is a short summary of the work to be presented in the article.
\begin{abstract}
This paper proposes a novel study on personality recognition using video data from different scenarios. Our goal is to jointly model nonverbal behavioral cues with contextual information for a robust, multi-scenario, personality recognition system. Therefore, we build a novel multi-stream Convolutional Neural Network framework (CNN), which considers   multiple   sources   of   information. From a given scenario, we extract spatio-temporal motion descriptors from every individual in the scene, spatio-temporal motion descriptors encoding social group dynamics, and proxemics descriptors to encode the interaction with the surrounding context. All the proposed descriptors are mapped to the same feature space facilitating the overall learning effort. Experiments on two public datasets demonstrate the effectiveness of jointly modeling the mutual Person-Context information, outperforming the state-of-the art-results for personality recognition in two different scenarios. Lastly, we present CNN class activation maps for each personality trait, shedding light on behavioral patterns linked with personality attributes.
\end{abstract}

% <concept_id>10003120.10003121</concept_id>
% <concept_desc>Human-centered computing~Human computer interaction (HCI)</concept_desc>

% \ccsdesc[500]{Human-centered computing~Human computer interaction (HCI)}

% %
% % Keywords. The author(s) should pick words that accurately describe the work being
% % presented. Separate the keywords with commas.
% \keywords{Personality recognition, CNN networks, social behaviors analysis, nonsocial behavior analysis}

%
% A "teaser" image appears between the author and affiliation information and the body 
% of the document, and typically spans the page. 
%%  \includegraphics[width=\textwidth]{sampleteaser}
%%  \caption{Seattle Mariners at Spring Training, 2010.}
%%  \Description{Enjoying the baseball game from the third-base seats. Ichiro Suzuki preparing to bat.}

% This command processes the author and affiliation and title information and builds
% the first part of the formatted document.
\maketitle

\section{Introduction}

Human personality and its influence on behaviors have been studied extensively in the field of psychology \cite{rammstedt2007measuring}. Independently from the situation or the geographical location \cite{schmitt2007geographic}, humans tend to show consistent behavioral patterns, defined by the psychologists as personality traits. 
The Big-Five trait model \cite{mccrae1992introduction} is the most popular one, dividing human behaviors in five general dimensions: Extroversion (i.e. talkative, outgoing), Agreeableness (i.e. considerate, forgiving), Conscientiousness (i.e. efficient, motivated), Neuroticism (i.e. tense, full of worries), and Openness to Experience (i.e. inventive, imaginative). With the recent rise in data usage and data analysis, there has been an increasing interest in the automatic assessment of personality traits using sensory data. 
%%%
%This research path is especially relevant for the ambient assisted living (AAL) or video surveillance domains. 
%%%
% psychological theories and machine learning are combined 
\\
Personality computing \cite{vinciarelli2014survey} is the field in which theories coming from the areas of machine learning and psychology merge to create computational models for personality understanding. However, given the uncertain delineations of human behaviors in different situations, it is extremely challenging to build general systems, which would work in diverse scenarios. As a consequence, recent personality computing systems have mostly focused on context-specific problems (e.g. job screening interviews \cite{ponce2016chalearn}, activities of daily living \cite{dotti2018behavior}, and work meetings \cite{zen2010space}). In contrast, in this study, we focus on building a novel and general architecture for personality  recognition suitable for different contexts. In computer vision, contextual information has been shown to improve several challenging tasks such as action recognition \cite{marszalek2009actions} and social scene understanding \cite{bagautdinov2017social}. Building on these findings, we propose a new framework, which maps the mutual relation between individual behaviors and contextual information to personality labels.

\begin{figure*}[!t]
\centering
\includegraphics[height=7cm,width=14cm]{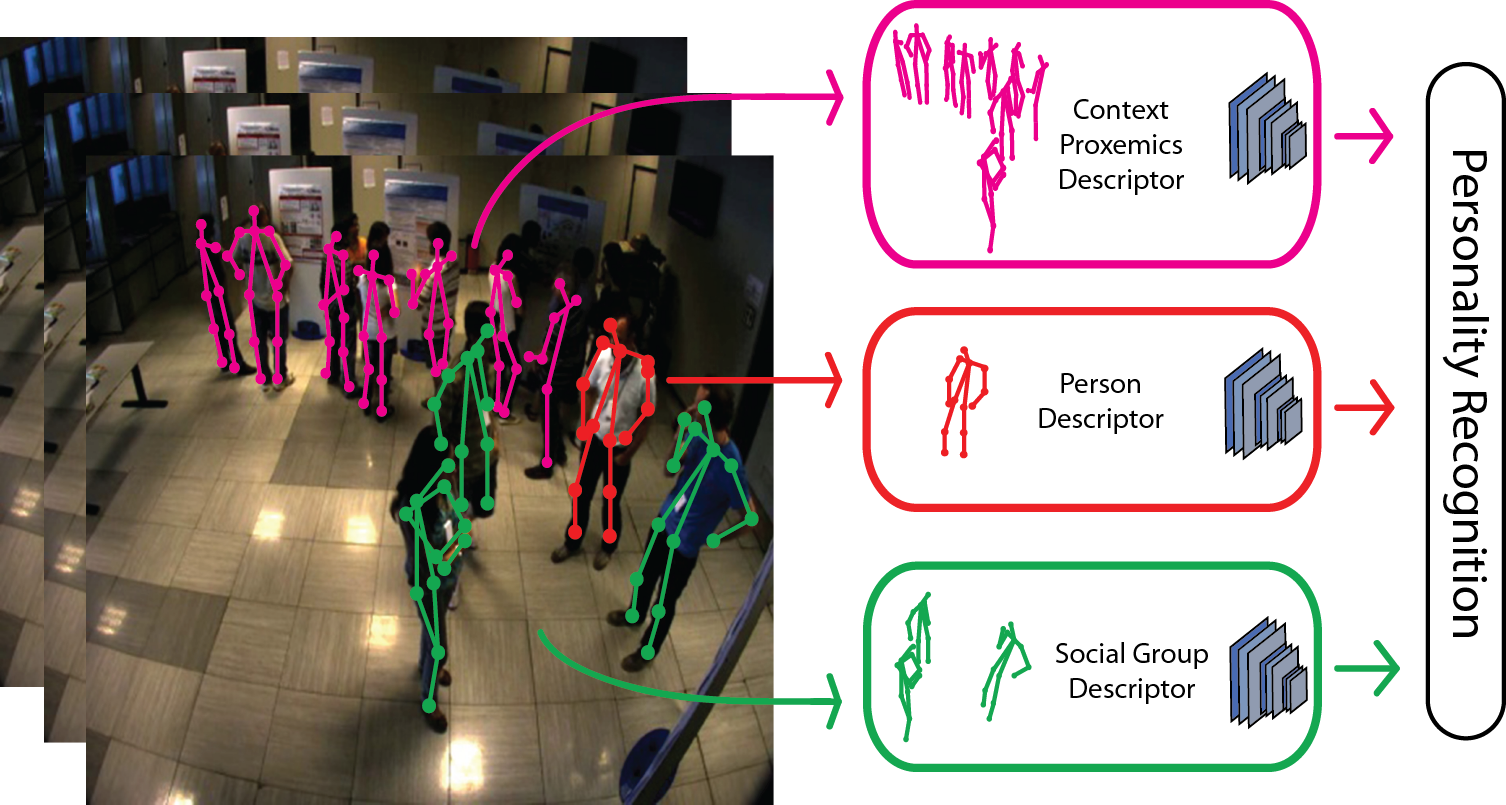}

\caption{High level description of the proposed model. Individual motion descriptors as well as two Context descriptors are learned in a novel CNN framework for personality recognition. Individual motion descriptors (red color) indicate the engagement level of every person in the scene, social group descriptors (green color) indicate the engagement level of individuals in conversational groups, and finally, the context proxemics descriptors (purple color) indicate the global attitude of each individual in respect to the others.}
\label{fig:high_level_img}
\end{figure*}

Imagine a social scenario in which different interpersonal styles influence the group interaction, for example, individuals that strive to be \textit{the center of attention} will try to actively capture the attention of the rest of the group, resulting in the group acting more passively \cite{snyder1986personality}. By combining information at the individual level with information at the context level, a robust semantic understanding of the situation is perceived. Even if individuals are in a nonsocial scenario (e.g. when an individual is alone at home), it has been shown that people engage with contextual objects as they would engage with other humans (i.e. Anthropomorphism) \cite{epley2007seeing}. Therefore, by examining the interaction between human behaviors and the surrounding scene, important information about human personality can be extracted. 

In the field of smart monitoring systems, the affective/emotional aspects of human behaviors are often omitted \cite{cristani2010socially}, whereas, it has been shown that actions and movements made by the individuals are influenced by their affective state, as well as their personality attributes \cite{yi2016pedestrian}. For example, authors in \cite{yi2016pedestrian} show that the pedestrians' walking behaviors in a public environment can be described by affective attributes. Pedestrians are considered aggressive if their walking path is not influenced by the surrounding crowd, whereas pedestrians are considered conservative if their paths are modified to avoid contact with the crowd. Similarly, authors in \cite{cristani2010socially} introduce the use of personal space as a source of information for surveillance systems. By considering mutual distances and mutual orientations, socially relevant material can be derived for assessing unexpected or abnormal situations. In this paper, we build on these findings to create a behavioral monitoring framework that considers the users' personality in different scenarios, and in this way, we aim to fulfill one of the most important systems design principle: ``know the user'' \cite{hansen1971user}. There are several ways in which users' information can be gathered through an interactive system, the standard way is the ``active interaction'', where the user is actively engaged with the system, for instance by pushing buttons or performing some actions. In the last years, with the advance of the technological power as well as sensors precision, studies have been presented in which the user is not directly engaged with the system, but through measurements and observations, the system is able to predict what the user wants. In this direction, we believe that automatically recognizing users' personality through ``passive interactions'' can improve the user's experience as well as can increase the system acceptance \cite{achten2013buildings}. For example, the user does not have to actively declare his/her current affective state, mood, or personality to the system, but the system aims to understand and subsequently make actions adapted to the user's personality \cite{mairesse2010towards, tapus2007hands, wang_agents2014}.  Therefore, by making automatic personality recognition systems more accurate and robust to different contexts, we are improving the system capacity to interact and adapt more like a ``human'' (i.e. perceiving more subtle human characteristics).

%We believe that automatically recognizing users' personality is of great help to enhance the field of Human-Computer Interaction (HCI), in which, most of the users' information is still gathered manually \cite{xu2015classroom}.

%For example, deploying an interactive intelligent system, able to assess the users' personality, would lead to a more pleasant and improved experience for the user, while increasing the system acceptance and efficiency. Examples of systems embodied with a certain personality type are already present on the market \cite{alexa2018}, or have been investigated from a research point of view in projects proposing robots in work environments \cite{you2018human}, for supporting children \cite{celiktutan2017multimodal}, or smart assistants reacting based on the user's personality and affective state \cite{wang_agents2014}. 

In the last few years, deep neural network (DNN) architectures achieved reliable results in the field of Personality Computing, using video \cite{dotti2018behavior}, audio \cite{lin2018using}, as well as multimodal data \cite{wei2018deep, guccluturk2017multimodal}. However, despite the growing effort, many challenges still remain to be tackled. First of all, datasets providing personality labels are still limited, resulting in ad-hoc methods for specific contexts \cite{alameda2016salsa, dotti2018behavior}, and, secondly, findings about personality are often too restricted to a research field (i.e. either psychology or computer vision). Thus, interdisciplinary research providing both quantitative, as well as qualitative results, are crucial towards more robust behavior understanding systems. 
\\
In this paper, we propose a novel architecture based on Convolutional Neural Networks (CNNs) \cite{simonyan2014two} for personality recognition, by studying the interaction between Person and Context information in a general manner, both in a social and a nonsocial context. In Fig. \ref{fig:high_level_img}, we show the high-level description of the proposed system. Apart from the Person motion Descriptor (red color), computed for every individual in the scene, context information is extracted in different levels of granularity. On the one hand, we compute the Social Group Descriptor (green color) on individuals engaging in close social interactions, where, by social interactions we imply the mutual interaction between two or more people \cite{cristani2011social}. On the other hand, we compute the Context Proxemics Descriptor (purple color), considering the interpersonal distance between all the individuals in the scene. Additionally, we show that our Context Proxemics descriptor is general enough to be applied also in a nonsocial scenario, encoding human-object interactions. As a result, both the local, as well as the global behavioral cues of an individual are mapped to personality labels.

Our contributions are as follows: 
\\
First, we create a novel, end-to-end, multi-stream CNN framework to analyze individual, as well as context information, for personality recognition using video data. This method is robust to different types of data (e.g. RGB or Depth image), different types of camera settings (e.g. fps and resolution), as well as different scenarios (e.g. social as well as nonsocial situations).
\\
Second, in addition to individual motion descriptors \cite{ke2017new}, that are transferred from the activity recognition field, novel CNN descriptors are proposed to encode the surrounding context in different scenarios. Transforming different sources of information (e.g. Person-Context) into CNN descriptors with the same backbone structure has the great advantage of facilitating the discovery of common latent representations. %Additionally, a novel pooling strategy is added, to encode the social group interaction. 
\\
Third, to demonstrate the generalizability of our algorithm, experiments are carried out on two public datasets. Results show that our Person-Context model outperforms the state-of-the-art methods. Additionally, to demonstrate the robustness of the learned personality patterns, we evaluate our framework using two different sets of personality classes as training labels (e.g. personality traits \cite{mccrae1992introduction}, as well as personality types \cite{block2014role}).
\\
Fourth, by visualizing the CNN activation map for both high and low scores of each personality trait, we pose the following questions: 1) \textit{Is the relation between Person-Context dynamics reflected in the traits?} and 2) \textit{Do the dynamics correspond to the trait attributes defined by psychologists?}. Our qualitative results provide new insights into the interaction between context and individuals' behavioral cues for behavior and personality understanding.

\section{Related works}

%Personality Computing aims to bridge the gap between human behavior understanding and computational systems. 
In this section, we describe works that are closely related to ours, in both the fields of computer vision and personality computing.

\subsection{Skeleton motion features}

Human motion recognition is one of the most important challenges in the computer vision community, because of its great applicability in several real-world challenges, such as video surveillance and activity recognition \cite{wang_survey2018}. In the past years, a multitude of methods have been proposed spanning across RGB-based features, Depth-based features, RGB-Depth-based features and skeleton-based features, combined with a plethora of Deep Neural Network (DNN) models \cite{wang_survey2018}. Since in this paper we utilize skeleton-based features, we will mainly focus on this type of information.

Skeleton joint estimation has shown promising results in the tasks of pose estimation 
\cite{cao2016realtime, zhang2018poseflow} and activity recognition \cite{ke2017new, liu2017skepxels}. However, by using only coordinates from skeleton joints, a considerable portion of RGB information is discarded, limiting the applicability of CNN methods. To overcome this issue, in the last years, novel strategies have been proposed to create feature representations similar to images  \cite{dotti2018behavior, ke2017new, liu2017skepxels}. Authors in  \cite{ke2017new} proposed a spatio-temporal representation of skeleton joints called ``image clips'' for action recognition. These features allowed the use of the VGG pre-trained model \cite{simonyan2014very} for feature extraction. Similarly, in \cite{liu2017skepxels}, a skeleton image of varying dimensionality is constructed to take advantage of the 2D CNN receptive field. These approaches brought a new way of encoding the skeleton joints, showing promising results for action recognition. 

The multitude of methods, as well as the important amount of available data proposed for human motion recognition opens the way to transfer the acquired knowledge to personality computing. For this reason, in this paper, we chose to explore the relation between skeleton motion features and personality labels, due to their compact representation, as well as robustness to occlusions or inter-personal variations. We show that our approach, inspired by \cite{ke2017new}, and optimized for a CNN architecture, outperforms the state-of-the-art for personality recognition.

%\textbf{Personality recognition.}
\subsection{Personality recognition}

Several datasets for personality recognition have been proposed in the recent years, both in controlled scenarios \cite{celiktutan2017multimodal, ponce2016chalearn}, and real life scenarios \cite{alameda2016salsa, dotti2018behavior}. These datasets are recorded using video data \cite{dotti2018behavior}, multimodal data ( video and audio \cite{ponce2016chalearn} or video and wearable sensors \cite{alameda2016salsa}), as well as  neuro-physiological data \cite{MirandaCorrea2018AMIGOSAD}.

A popular approach towards personality computing is to link face and audio features to personality traits \cite{wei2018deep}, and recently, authors in \cite{persEmoNet2018} showed that facial and audio features can be used to link personality and emotion cues.
However, systems that apply face analysis are not robust to camera positions, as well as to privacy restrictions \cite{bowyer2004face}. To overcome this obstacle, it has been shown that body motion patterns are also a major indicator of personality traits \cite{alameda2016salsa, dotti2018behavior}.  Authors in \cite{alameda2016salsa} found that high extroversion trait is highly correlated with body postures predisposed towards face to face conversations, while \cite{koppensteiner2013motion} showed that upper body motion of public speakers is a good predictor for personality. Body pose features in combination with Deep Boltzmann Machines (DBMs) proved their efficiency at discovering the emergent leadership in a small group \cite{beyan2018investigation}. In \cite{dotti2018behavior}, posture dynamics are learned using an Autoencoder-LSTM architecture, discovering motion patterns for personality labels. In this direction, we propose a model for personality recognition, to discover discriminative nonverbal behavioral cues for each personality trait.

%\textbf{Social and nonsocial interaction.} 
\subsection{Social and nonsocial interaction}
The analysis of social interactions is an important challenge for several applications, like video surveillance \cite{cristani2011towards} and activity recognition \cite{biswas2018structural}. Social interactions are rich in information, revealing characteristics of both single individuals as well as groups. In \cite{biswas2018structural}, authors proposed an LSTM based system to model, at the same time, individual and group activities. In \cite{wang2017recurrent}, an LSTM system is proposed to model intra-group dynamics (person within a group) and inter-group (group to group) interactions. When it comes to nonsocial interaction, human-object interaction has been studied by several authors. In \cite{marszalek2009actions}, the correlation between actions and context has been explored, showing how actions are constrained by certain scenes. In \cite{yao2012recognizing}, the mutual context between human poses and objects has been modeled for Human-Object Interactions. Following these studies, this paper studies the relation between Person-Context interaction dynamics and personality behavioral patterns.

\section{Person-Context Framework}

In this paper, we jointly model the nonverbal behavioral cues from single individuals, with surrounding context information, for personality recognition using video data. Theoretical models such as the Laban Movement Analysis (LMA) \cite{hutchinson1954labanotation} conceptualize the relation between human motion and the surrounding space. Although this model was proposed initially to describe dance movements, authors of \cite{roudposhti2013probabilistic} and \cite{roudposhti2016probabilistic} adopted certain concepts for Human to Human and Human-Robot Interactions understanding. Moreover, while psychological studies have demonstrated the tight relation between contextual information and personality patterns \cite{lau2008interplay}, works considering both sources of information are quite limited in the computer vision community. Therefore, inspired by these interdisciplinary research themes, we propose an automatic personality recognition system using a person-context deep model.

\begin{figure*}[!t]
\centering
\includegraphics[height=6cm,width=14cm]{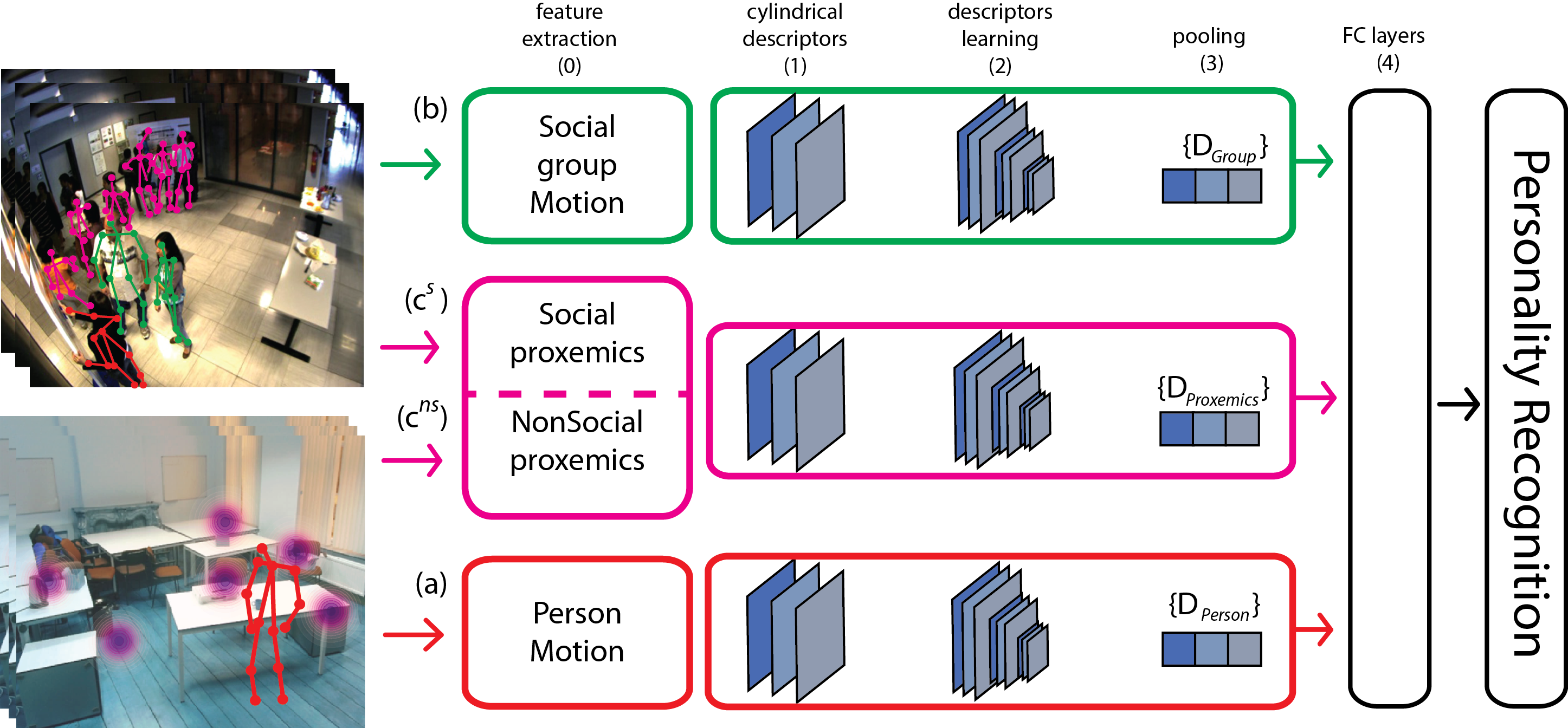}

\caption{Framework architecture. The Person-Context streams (a, b, c) are extracted from video sequences (0) and processed separately (1, 2, 3). The output descriptors are fused in FC layers (4) and finally, a softmax layer is employed for personality recognition.}
\label{fig:architecture}
\end{figure*}

\subsection{Architecture}

The proposed architecture is shown in Fig. \ref{fig:architecture}. For every frame $I^{t}$ in a given sequence, we first run a pose estimation algorithm \cite{cao2016realtime}, with the role of defining semantic components of the scene (Fig. \ref{fig:architecture} (a), (b) and (c)). In case some of the body joints are not detected correctly, we estimate their positions using the neighbours' joints coordinates. As the algorithm proposed by \cite{cao2016realtime} does not track the detected objects, a frame by frame tracking algorithm is added \cite{opencv_library}. Given the detected skeletons $p_{subjs}(subjs \in {1,..,n})$ where $n$ is the total number of subjects in a social $sc^s$ or nonsocial scene $sc^{ns}$, we aim to map Person as well as Context information to personality labels. One disadvantage of jointly modeling different sources of data is that their intrinsic structure is too diverse, making the learning process more complex. Differently, one key aspect of the proposed method is to use the same backbone structure (i.e. cylindrical coordinates descriptors, Fig. \ref{fig:architecture} (1)), to describe the various types of information. In this way, different feature modalities are mapped onto the same feature space, allowing the creation of the same network structure (Fig. \ref{fig:architecture} (2) and Fig. \ref{fig:architecture} (3)). Note, the extracted components (Fig. \ref{fig:architecture} (a), (b) and (c)) are general enough to deal with nonsocial scenes $sc^{ns}$, where, the Social Group Interaction (Fig. \ref{fig:architecture} (b)) will be set to $0$, and the Context Proxemics Descriptor (Fig. \ref{fig:architecture} ($c^{ns}$)) will encode interactions with semantic objects in the scene. Finally, Fully-Connected Layers (FC) are adopted to fuse the features extracted by the distinct streams, and a Softmax layer is added for the personality recognition task. As follows, detailed explanation of every component is provided: the Person descriptor is described in Sec. \ref{sec:person_stream}, then, novel Context CNN descriptors are presented in Sec. \ref{sec:social_interaction} and Sec. \ref{sec:context_prox}. The learning component is introduced in Sec. \ref{sec:cnn_learning}, and, finally, quantitative as well as qualitative results are reported in Sec. \ref{sec:experimets}, Sec. \ref{sec:qual_res} and Sec. \ref{sec:patterns} respectively.

\subsection{Person motion}
\label{sec:person_stream}

Nonverbal behavioral cues have been studied extensively as personality patterns descriptors \cite{dotti2018behavior, koppensteiner2013motion}. In this section, skeleton pose information, extracted using \cite{cao2016realtime}, is employed to create the ``Person'' descriptor (Fig. \ref{fig:descriptors}(a, $0$)), in the proposed Person-Context framework. Inspired by \cite{ke2017new}, we create temporal ``skeleton clips'', to describe the relative motion of each skeleton joint $J^{joints}$, where $joints$ is the number of skeleton joints detected. In particular, for every frame in a sequence $t$, the relative position of all skeleton joints is computed with respect to the selected reference joints ${Jr}^{ref}$, where $ref$ denotes the number of reference joints. Similar to \cite{ke2017new}, we set $ref=4$, resulting in $ref$ matrices of size $t \times J^{joints-1}$.  

To make our descriptor more compact, unlike \cite{ke2017new}, the computed distances are stacked vertically (Fig. \ref{fig:descriptors}(a, $1$)), to form a final matrix that describes along the vertical axis, the temporal evolution of the joints in a sequence $t$, and along the horizontal axis, the relative distance of the skeleton joints in one frame. Note that, the relative distances are computed for each coordinate independently, resulting in a 3D matrix. Since the joints positions are declared in the Cartesian coordinates, following \cite{ke2017new} and \cite{weinland2006free}, we transform the 3D temporal matrix into cylindrical coordinates (Fig. \ref{fig:descriptors}(a, $1$)). Authors in \cite{weinland2006free}, created motion features for action recognition using cylindrical coordinates to obtain features invariant to location, orientation and scale. This is particularly important in unconstrained multi-camera datasets, like the Salsa dataset \cite{alameda2016salsa}, in which the scene is crowded, and occlusions make the skeleton detection challenging. \\

\begin{equation}
\label{eq:cyl_co}
    (\rho, \theta, z) = (\sqrt{x^2+y^2}, \tan^{-1}\left(\frac{y}{x}\right), y_1-y_2)
\end{equation}

Cylindrical coordinates (Eq. \ref{eq:cyl_co}) are composed by three terms: $\rho$,  $\theta$, and $z$. The term $\rho$ is defined as the euclidean distance between the reference joints ${Jr}^{ref}$ and the observed joints $J^{joints}$, the term $\theta$ is defined as the angle between the reference joints ${Jr}^{ref}$ and the observed joints $J^{joints}$, and the term $z$ is the difference considering only the vertical axis between the reference joints ${Jr}^{ref}$ and the observed joints $J^{joints}$. Finally, the three descriptors values are converted between $0-255$, suitable for a CNN architecture. 

\subsection{Social Group Motion}
\label{sec:social_interaction}

Given a social scene $sc^{s}$, the modeling of social group interactions \cite{cristani2010socially} is a challenging task for behavior understanding. Since, interaction dynamics are affected by the behaviors of single individuals involved in the group, the understanding of each personality is of great importance to unfold the social group evolution. For example, an extroverted person will tend to be actively involved in the group, while an agreeable person will tend to be more passive \cite{alameda2016salsa}. In this section, we explain how the social group motion is encoded using the cylindrical descriptors (Eq. \ref{eq:cyl_co}).

The code provided by \cite{cristani2011social}, for social interaction discovery during social events, is used to detect the members of each social group in the scene. We define each member as $m_{M}^{group}$, where $M$ is the total number of individuals in a given social group $group$. Since social groups dynamics are affected by single individual's motion, we compute the motion dynamics (Sec. \ref{sec:person_stream}) for every member $m_{M}^{group}$. However, this introduces a new challenge: groups vary the number of members frequently, resulting in descriptors of different sizes. Therefore, a pooling strategy is adopted to encode the social motion dynamics of each group in a compact representation. This strategy ensures that the obtained features are more robust to the effect of group dynamics variation, while still preserving important data patterns.

\subsubsection{Social Motion Pooling}

In order to integrate the motion dynamics of $m_{M}^{group}$, we focus on the two most conventional pooling operations, max pooling and average pooling. Choosing the right pooling strategy is important to ``downstream''  the  descriptors to a fixed dimension independently from the social group size, as well as encoding the group motion dynamics.
 
Pooling operations are performed on the three cylindrical descriptors independently $(\rho, \theta, z)$. Average pooling is defined as $f_{ave}(x) = \frac{1}{M} \sum_{m=1}^{M} \sum_{i=1}^{N} x_{m,i}$, and max pooling is defined as \\
$f_{max}(x) = \max_{m,i} x_{m,i}$, where, the vector $x$ contains the activation values from pooling regions $i=[1, \dots, N]$, computed on every member of the social group $m= [1, \dots, M]^{group}$. 

An exploratory experiment is conducted to test the performance variation when using the described pooling strategies on the Salsa dataset poster session \cite{alameda2016salsa}. In particular, we aim to analyze the information carried by the social group motion descriptor in the task of personality recognition. In Fig. \ref{fig:avg_max_pool}, accuracy results for the classification of the big 5 personality traits, using the described pooling strategies are visualized.  The results indicate that average pooling constantly reaches higher accuracy, and therefore is selected as the social group pooling strategy in the rest of the experiments. However, it is interesting to notice that the performances are not statistically significant (indicated by the asterisks), when predicting two personality traits: Conscientiousness and Neuroticism.     

Average motion pooling encodes the motion from all the members in the social group, whereas max pooling encodes the highest motion value in the group. As individuals with high Conscientiousness or Neuroticism traits are less likely to stand out or influence the social group, the average social group motion dynamics may acquire more importance than the individual motion, and therefore confuse the classifier decision.

\begin{figure}[!t]
\centering
\includegraphics[height=9cm,width=13cm]{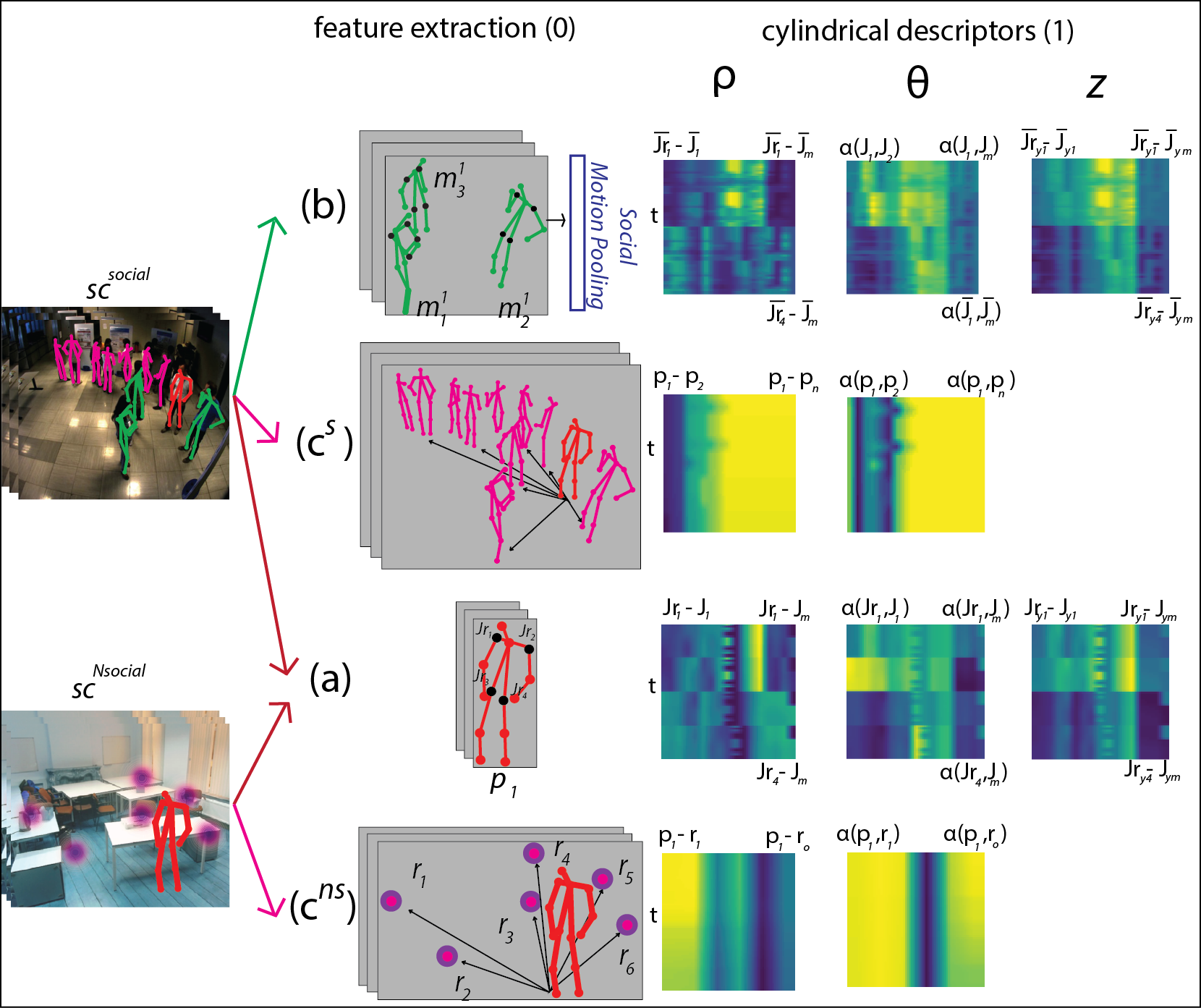}

\caption{CNN descriptors of the proposed framework: (a) Person motion descriptor, (b) Social group motion, ($c^{s}$) Social proxemics, and ($c^{ns}$) Nonsocial proxemics. The descriptors values $(0-255)$ range in color from blue (low distances between skeleton joints) to yellow (large distances between skeleton joints).}
\label{fig:descriptors}
\end{figure}

\subsection{Context proxemics}
\label{sec:context_prox}

Human behaviors are often correlated with the environment they are performed in, hence, in this section, proxemics descriptors encoding the interaction with different contextual information are introduced depending on the analyzed scenario. For example, during a social event, it is likely that individuals show social behaviors, whereas, when individuals are at home alone, it is likely that behaviors are related with objects in the scene. In the next sections, two strategies are proposed towards encoding proxemics to different entities.
%(i.e. interpersonal distance in a social scenario, and  distance to semantic objects in a nonsocial scenario).

%\begin{figure}[!t]
%\centering
%\includegraphics[height=6cm,width=10cm]{images/average_max_pooling_pred.png}

%\caption{Exploratory experiment investigating the recognition accuracy using two different pooling strategies encoding the social groups' motion. Average pooling demonstrates better performances over all the big 5 traits, however the results are not significant over two traits: Conscientiousness and Neuroticism. P values less than 0.001 are summarized with three asterisks, P values less than 0.01 are summarized with two asterisks, and finally, no asterisks show not significant p values. }
%\label{fig:avg_max_pool}
%\end{figure}

\subsubsection{Social Proxemics}
\label{sec:social_prox}

Several works highlighted the correlation between interpersonal distances (i.e. proxemics) and personality traits \cite{zen2010space} , \cite{cristani2011social} in social scenarios. Thus, in this section, a proxemics descriptor encoding the global position of $p_{subjs}$ in respect to all the other individuals in the scene, is computed.

In the analyzed scene $sc^{s}$, proxemics is intended as the way people use their personal space in relation to other individuals. The process is illustrated in Fig. \ref{fig:descriptors} ($c^{s}, 0$), where we compute $\rho$ and $\theta$ descriptors between $p_{subjs}$ and the rest of the individuals using Eq. \ref{eq:cyl_co}, for a frame sequence of length $t$. To avoid dimensionality discrepancy due to the different number of individuals in the scene, we always consider the maximum number $n$ of subjects for a given dataset. Note that the $z$ descriptor is not considered, as the interpersonal distance does not occur on the vertical axis. Thus, two descriptors of size $t \times n-1$ are obtained, and to be consistent with the other generated features, they are transformed into grey images by scaling the values between $0-255$ (Fig. \ref{fig:descriptors} ($c^{s}, 1$).

These descriptors have two advantages: Firstly, the global position of every subject with respect to all the other subjects in the scene is represented, reflecting behaviors that have been already correlated with personality display \cite{alameda2016salsa}. For example, individuals who like to be engaged in group conversations (Agreeableness personality trait) will have a  higher number of neighbours, while on the other hand, shy individuals will have fewer neighbours. Secondly, due to the common cylindrical coordinates structure, these descriptors can be fused with the other generated features, optimizing the learning capacity of our CNN architecture.

\subsubsection{Nonsocial proxemics}
\label{sec:nonsocial_prox}

As humans are by nature social beings, personality displays have been well explored during social interaction. However, as reported by \cite{dotti2018behavior}, few efforts have been made towards the understanding of personality in a nonsocial context. This problem has been shown to be important for applications in domains like Ambient Assisted Living (AAL) and Smart Homes, where it often occurs that individuals are spending their time at home alone. Thus, in this section, we aim to learn the relation between human behaviors and the surrounding nonsocial context, defining proxemics as the way people use their personal space in relation to objects (Fig. \ref{fig:descriptors} ($c^{ns}$)). 

We build on the philosophical and psychological theorem which states that people engage with objects as they engage with other humans (e.g. Anthropomorphism \cite{epley2007seeing}). In this section, a novel descriptor is proposed to capture the way individuals engage with the scene. As shown by \cite{epley2007seeing}, especially in case of ``lonely'' situations, subjects are more likely to be subjected to anthropomorphism with nonhuman elements like objects, robots etc. 

Since no a priori information is given about the scenario, firstly, semantic regions of the scene are discovered in an unsupervised way and, secondly, we compute a temporal descriptor optimized for a CNN architecture, encoding Person-Context interactions.
Spatial information is extracted by dividing the scene into fined-grained patches. In every patch, we assume that the arm motions are the most important joints for person-context interaction location. Therefore, we quantize motion, as well as orientation information of the arms in a histogram of oriented tracklets descriptor \cite{mousavi2015analyzing}. However, movements such as walking generate motion from the arms but are not interesting for our Person-Context Interaction investigation. To avoid this, the average body motion is subtracted from the arm motion. To locate the regions of the scene where the interaction occurs in an unsupervised way, Gaussian Mixture Model (GMM) is applied \cite{banfield1993model}. GMM clustering is a method which attempts to find a mixture of multidimensional Gaussian probability distributions that best model the input data. Because GMM is a probabilistic model, we can treat the clustering task as ``soft'' assignment, being more robust to outliers and noisy data. In Fig. \ref{fig:descriptors} ($c^{ns}$), the discovered semantic regions $r^{regions}$ are shown, where $regions = 6$.

To extract descriptors that reflect the engagement of individuals in the generated semantic regions, we adopt the same strategy as in Sec. \ref{sec:social_prox}, computing $\rho$ and $\theta$ (Eq. \ref{eq:cyl_co}) descriptors between $p_{subjs}$ and the generated regions, for every frame sequence $t$ (Fig. \ref{fig:descriptors} ($c^{ns}, 0$). Finally, two descriptors of size $t \times regions$ are obtained, and transformed into grey images by scaling the values between $0-255$ (Fig. \ref{fig:descriptors} ($c^{ns}, 1$).

\section{Person-Context Interaction Learning using a CNN architecture}
\label{sec:cnn_learning}

The integration of multiple feature representations, is a challenging task due to the heterogeneity of their distributions \cite{zhao2015heterogeneous}. However, CNN architectures showed great ability in discovering common latent representations. Therefore, in this paper, a CNN framework is proposed for the fusion and learning of features extracted from multiple sources: individuals' motion (Sec. \ref{sec:person_stream}), social group motion (Sec. \ref{sec:social_interaction}) and context proxemics (Sec. \ref{sec:context_prox}).  In order to reduce the structural difference of different distributions, a common feature representation based on cylindrical coordinates is created (Fig. \ref{fig:architecture}(1)). This strategy allows the CNN model to use the same parameters (e.g. number of hidden units, pooling layers etc.) for all the feature representations, decreasing the duration and complexity of the training phase (Fig. \ref{fig:architecture}(2), Fig. \ref{fig:architecture}(3), Fig. \ref{fig:architecture}(4)).

\begin{figure}[ht]
\centering
\subfigure[]{\label{fig:avg_max_pool}
\includegraphics[height=4.5cm,width=6.8cm]{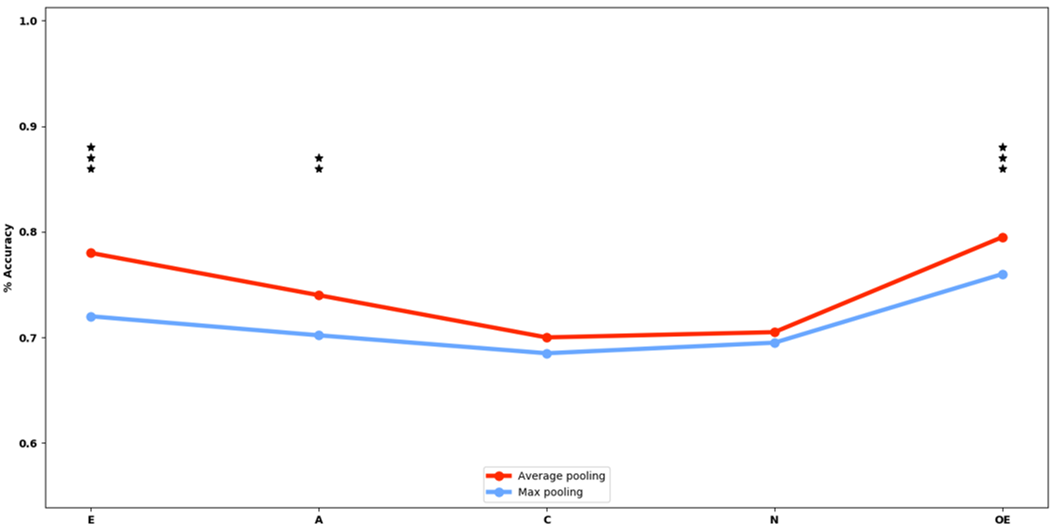}}
\subfigure[]{\label{fig:fusion_score}
\includegraphics[height=4.5cm,width=6.8cm]{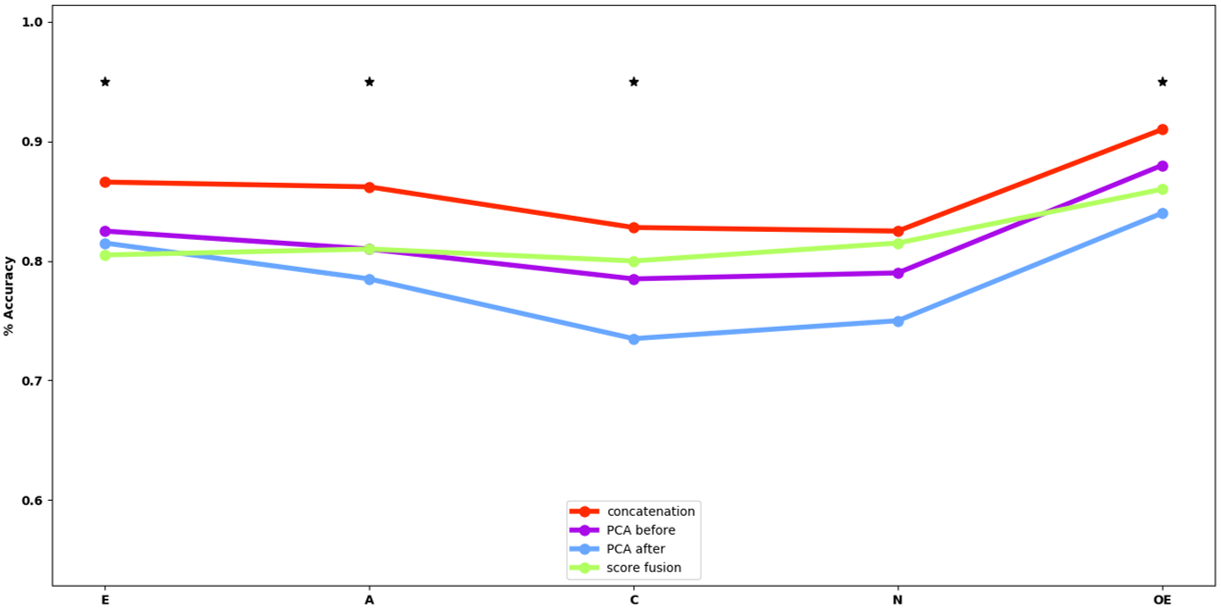}}

\caption{(a) Exploratory experiment investigating the recognition accuracy using two different pooling strategies, encoding the social groups' motion. Average pooling demonstrates better performances over all the big 5 traits, however the results are not significant over two traits: Conscientiousness and Neuroticism. The p-values less than 0.001 are summarized with three asterisks, p-values less than 0.01 are summarized with two asterisks, and finally, no asterisks show not significant p-values. (b) Exploratory experiment investigating the recognition accuracy using different fusion methods. Feature concatenation shows better results than the other methods, being significantly superior on four out of five traits. The p-values less than 0.05 are summarized using one asterisk.}
\end{figure}

\subsection{Feature Learning}

The temporal resolution of each descriptor varies depending on the fps of the analyzed dataset ($t \in {15,30}$ frames), while, in order to be fed to the CNN network, their final size is scaled to ($68\times68\times3$). This dimensionality is chosen to preserve the mapping of the skeleton joints indexes on the original vector, with the rescaled descriptors. This is particularly useful, for example, for the visualization of the CNN layers (Sec. \ref{sec:patterns}). Given the CNN descriptors as input, the VGG19 model \cite{simonyan2014very} pre-trained on ImageNet \cite{russakovsky2015imagenet}, is used as a first step in the learning phase. Note that we utilize the VGG19 model only for feature extraction discarding the last 3 fully-connected layers. Early convolutional layers showed to learn more generic features, whereas, deeper layers are more influenced by the task they are trained for \cite{ke2017new}. Since our CNN representations are very different from the images contained in ImageNet, we extract the features from the convolutional layers $conv5_1$. We justify the adoption of this layer, as authors in \cite{ke2017new} applied it on similar image clips. Specifically, the generated cylindrical descriptors are fed as input to the VGG model. As output, from the $conv5_1$ we obtain feature representations of size $4 \times 4 \times 512$ ($512$ feature maps of size $4 \times 4$). 
Since the VGG feature extractor projected the Person-Context information in a lower-dimensional feature space, a pooling strategy is adopted to encode semantic spatio-motion features. Specifically, as in \cite{ke2017new}, we apply temporal mean pooling (TMP) to extract temporal information from the sequences.

After the pooling step, the local features, defined as $D_{pers}$, $D_{group}$, and $D_{prox}$ in Fig. \ref{fig:architecture} (3), are fused to form the input to two Fully-Connected Layers (FC) and the Softmax layer for the final classification. Between the two FC layers there is a rectified linear unit (ReLU) \cite{nair2010rectified} to introduce an additional non-linearity.

\subsubsection{Fusion}

The fusion of different input cues can improve the recognition performance, as it combines different sources of information that are relevant to discover personality patterns. In our framework, fusion can be performed as early fusion (feature fusion) or late fusion (decision fusion) \cite{snoek2005early}. 

As feature fusion, we implemented the following methods: 1) concatenation of the features coming from different modalities. 2) Principal Component Analysis (PCA) \cite{wold1987principal} was applied on the features extracted from different modalities, for dimensionality reduction, as well as to remove correlations between the features. Following \cite{alghowinem2016multimodal}, 98\% of variance was kept. 3) PCA was applied on the concatenated features. 
As decision fusion, the network was trained on the three descriptors (e.g. $D_{pers}$, $D_{group}$, and $D_{prox}$) independently and the recognition decision was fused (sum-rule) \cite{alghowinem2016multimodal}. Fig. \ref{fig:fusion_score} shows the personality traits recognition results using the described fusion methods on Salsa dataset poster session. The concatenation method exhibits the best results on all the big 5 traits, being significantly higher on four out of five traits (p-value less than 0.05 is summarized using one asterisk). Therefore, this method is used for the reminder of the experiments.

\section{Experiments}
\label{sec:experimets}

In order to examine the strength of our framework, in this section, we present personality recognition experiments on two public datasets (Sec. \ref{sec:dataset}). The choice of these two datasets justifies the evaluation of the system in two different scenes, where, depending on the scenario, the deep Person stream is combined with distinct Context streams (Fig. \ref{fig:descriptors}). The experiments are organized as follows: in Sec. \ref{sec:pers_rec_types}, we compare our work with state-of-the-art systems for personality recognition, and  in Sec. \ref{sec:pers_rec_traits}  we evaluate our framework using the Big-Five personality traits as labels, and finally, qualitative results are explained in Sec. \ref{sec:qual_res}.

\subsection{Datasets}
\label{sec:dataset}

The Salsa dataset \cite{alameda2016salsa} contains multimodal data from two social events (30 minutes each) in an unconstrained indoor scenario. Video data was recorded from 4 cameras placed at each corner of the room, and ground truth positions of the subjects' movements were provided every 45 frames. It consists of two parts, the first part was recorded during a poster presentation session, and the second one was recorded during the coffee break, where all the participants were allowed to freely interact with each other (this part is named cocktail party). The two parts contain the same participants and their personality scores were collected using the Big-Five personality questionnaire \cite{john1999big}. The personality in a nonsocial context \cite{dotti2018behavior} is a recent dataset, providing video data of 45 participants in an unconstrained indoor scenario. Every subject performed six tasks resembling Activities of Daily Living (ADL) and filled the short version of the Big-Five personality questionnaire \cite{rammstedt2007measuring}.

\subsection{Experimental Setup}

\begin{figure}[ht]
\centering
\subfigure[]{\label{fig:hierarchical_cl}
\includegraphics[height=4cm,width=6cm]{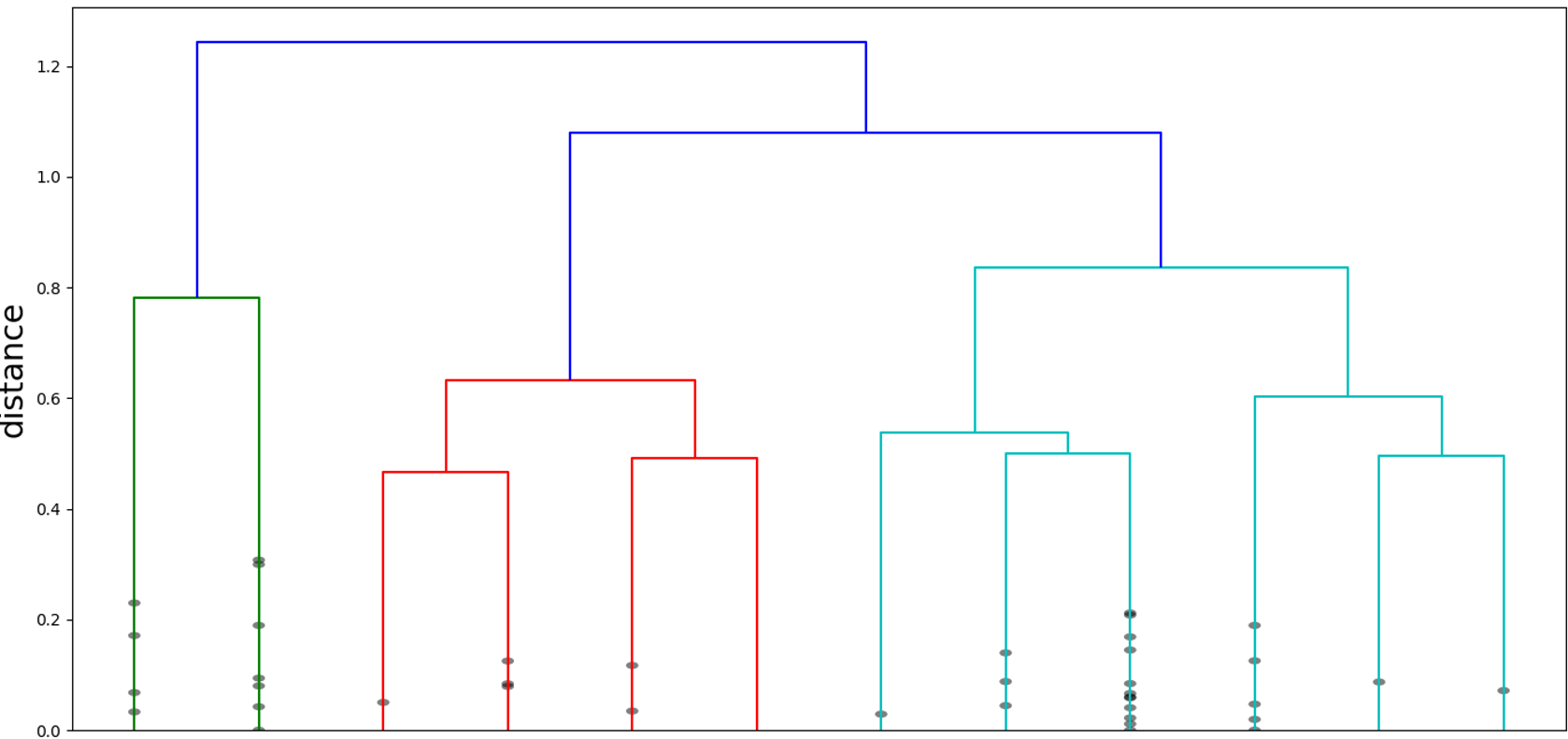}}
\subfigure[]{\label{fig:pers_types_dist}
\includegraphics[height=4cm,width=6cm]{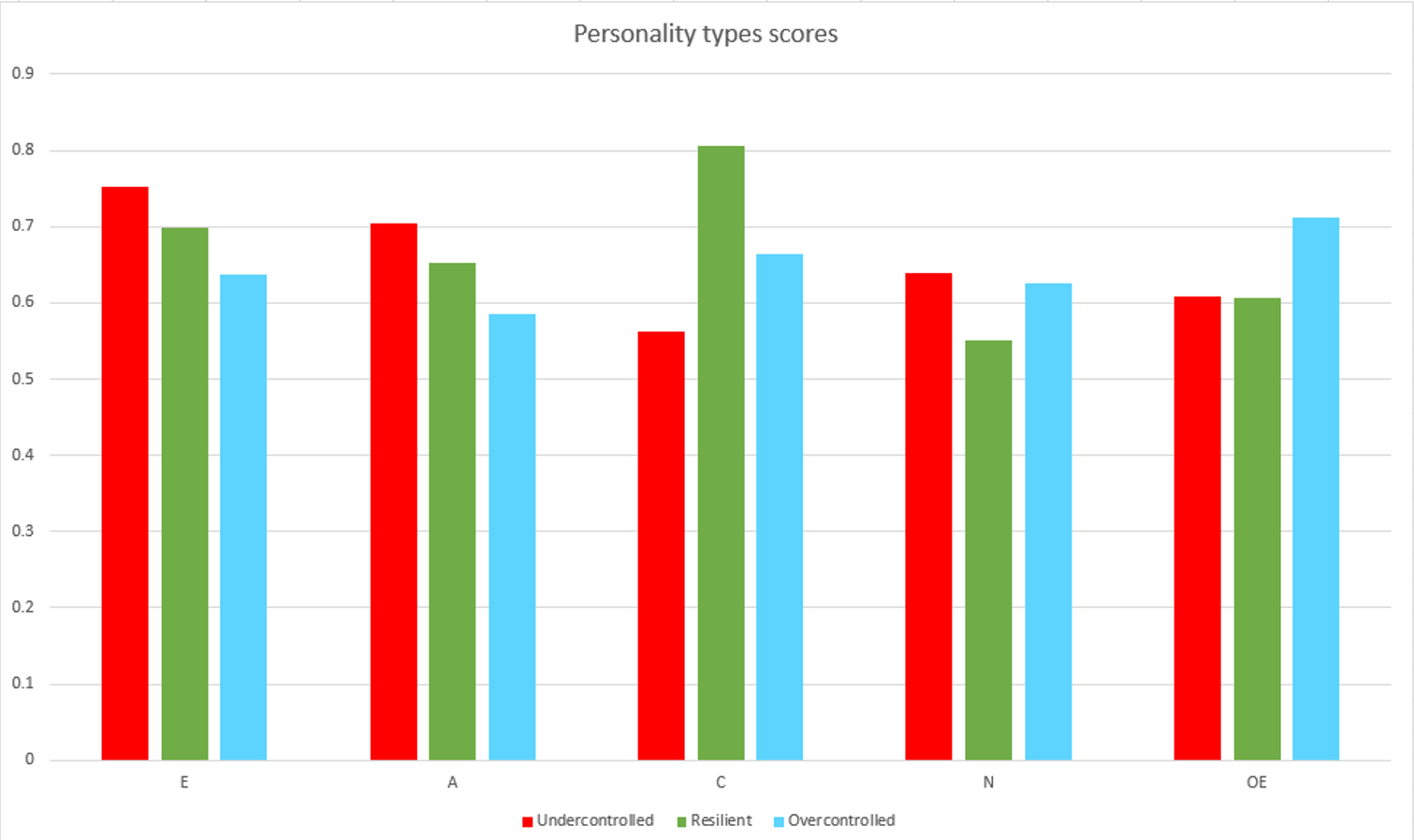}}
\caption{(a) Dendrogram of the hierarchical clustering applied on the normalized personality trait scores from the used datasets. Three main clades (coloured in green, red and azure) are found. (b) Average values for each personality trait in the discovered personality types (underconrolled personality type colored in red, resilient personality type colored in green, and overcontrolled personality type colored in azure).}
\end{figure}

Since the research in \cite{dotti2018behavior} is the most related to our work, the same experimental procedures are followed. In particular, the personality trait scores from the two datasets are normalized (between $0,\dots,1$), and a hierarchical clustering technique \cite{corpet1988multiple} is applied to explore the traits configuration. Confirming the findings in \cite{dotti2018behavior}, $clusters=3$ are found. As shown in Fig. \ref{fig:hierarchical_cl}, the traits scores are grouped in three main clades (colored in green, red and azure). Then, in order to assign a semantic meaning to the discovered clusters, we compute the average score of each personality trait in the different clusters. 

In Fig. \ref{fig:pers_types_dist}, we show the obtained results, which are consistent with the psychological theory proposed by \cite{block2014role}, stating that the Big-5 personality traits can be organized in three major types: undercontrolled, overcontrolled and resilient \cite{dotti2018behavior}. In Fig.  \ref{fig:pers_types_dist}, resilient personality type (green color) shows high Extraversion score and the lowest score in Neuroticism, the undercontrolled type (red color) scores high in Extraversion as well as Neuroticism, and finally, the overcontrolled type (azure color) has the lowest score in Extraversion and scores high in Neuroticism. 

These new classes are used as labels in the personality recognition experiments. Moreover, regarding the classification task, using personality types labels, two testing approaches are carried out: ``K-cross validation'' where $K=10$ and ``leave-one-out'' (specifically, the data from one subject is left out from the training procedure and used only for testing).

\textbf{Ablation Study.} In order to test the contribution of every component in the proposed framework, different combinations are tested. 
For the Salsa dataset, we tested: $D_{pers}$, where only the Person motion stream is used. $D_{prox}+D_{group}$, where only the two context streams are used. $D_{pers}+D_{prox}$, where person motion (Sec. \ref{sec:person_stream}) is combined with social proxemics descriptors (Sec. \ref{sec:social_prox}), $D_{pers}+D_{group}$, in which the social group motion descriptors are used (Sec. \ref{sec:social_interaction}), and finally, $D_{pers}+D_{prox}+D_{group}$, where all the descriptors are fused. As for the nonsocial context, we tested: $D_{pers}$ stream alone, and $D_{pers}+D_{prox}$ encoding the person motion with the scene proxemics (Sec. \ref{sec:nonsocial_prox}).

\textbf{Baseline Methods.} We compare our framework against the three most related works. In particular, $LSTM_{cl}$, developed by \cite{dotti2018behavior} was initially proposed for personality recognition on the nonsocial dataset. $Clips+MTLN$ was proposed by \cite{ke2017new}, and it uses similar skeleton descriptors as input to a CNN framework called MTLN. Finally, EL-LMKL \mbox{\cite{beyan2018investigation}} was proposed for leadership recognition as well as extraversion trait recognition on a leadership Dataset \cite{sanchez2011nonverbal}. Please note that all the baselines were implemented by the authors, as to the best of their knowledge, there do not exist personality recognition results on the Salsa dataset.

\subsection{Personality types recognition}
\label{sec:pers_rec_types}

Personality recognition results obtained on the two parts of the Salsa dataset, as well as on the nonsocial dataset, are shown in Table \ref{table:salsa_types} and Table \ref{table:nonsocial_types} respectively. Overall, the proposed method is able to reach the best accuracy results in both datasets, demonstrating the benefit of fusing Person-Context information. 
%The results on the poster session part of Salsa dataset (Table \mbox{\ref{table:salsa_types}}) depicts that, although the fusion of all the context descriptors ($D_{pers}+D_{prox}+D_{group}$) obtains the highest accuracy, it does not improve significantly the result obtained using other descriptors (see Salsa Poster CV and Salsa Poster LOO results in Table \mbox{\ref{table:salsa_types}}). 
In Table \ref{table:salsa_types} we report the results on the two parts of Salsa dataset (poster and cockatil party sessions) for both testing approaches (Cross-validation and Leave-One-Out). For the results on the poster session part, the ablation study indicates that, although the fusion of all the context descriptors ($D_{pers}+D_{prox}+D_{group}$) obtains the highest accuracy, it does not improve significantly the result obtained using only $D_{pers}+D_{group}$. This may be explained by studying the scenario of the dataset, which displays interactions during a scientific poster session. During this type of sessions, individuals have to respect some spatial constrains, for example, if a poster presentation is already crowded, individuals cannot position themselves freely. Hence, in this situation, the proxemics descriptor does not always describe affective behaviors as explained by \cite{cristani2011social}. Additionally, note that $D_{pers}$ descriptor alone reaches higher accuracy than $D_{prox}+D_{group}$, supporting the explanation that proxemics features in this scenario does not show a clear link to personality patterns. On the other hand, the ablation study on the cocktail party part demonstrates the advantage of considering all the descriptors, where the three proposed descriptors ($D_{pers}+D_{prox}+D_{group}$) obtain the best accuracy, significantly improving all the other descriptors' combinations. As this part of the dataset depicts the participants freely interacting with each other, social behaviors are more natural and less constrained by social roles (i.e. poster presenter vs. audience). While natural behaviors may be more representative of a certain personality type, they are more complex to model, and as consequence, the overall recognition accuracy is lower. The complexity in this scenario affects all the parts of the framework, as free interaction and group formation create occlusions and clutter, challenging the tracking and motion modeling components. Finally, the results show that the fusion of all the descriptors is more robust to real-world cluttered scenarios than individual descriptors. \\
Furthermore, the statistical significance of the results over the two parts of Salsa dataset is computed in respect to the baselines (indicated by the number of asterisks in Table \ref{table:salsa_types}). As the poster session presents more static social groups that are not formed by free interactions but imposed by the rules of a poster session event, the combination of Context and Person information does not add a statistical improvement (p-value > $0.05$) in comparison to single motion information computed using \cite{ke2017new}. On the other hand, the results obtained on the second part of the dataset (i.e. cocktail party) show significance difference (p-value < $0.05$) demonstrating the benefit of fusing Person-Context information during social interactions. Finally, our experiments show that there exist many factors interfering with finding the connection between human behaviors and personality labels, such as within-person variability \cite{wang2018sensing} in different situations (i.e. poster session versus cocktail party session). For this reason, we also show the personality recognition results by averaging the two sessions ($78.52\%$ for cross-validation and $73\%$ for leave-one-out testing), which are significantly different than the baselines. Concerning the nonsocial dataset (Table \ref{table:nonsocial_types}), our method obtains the highest results, proving the benefit of combining Person-Context information also in a nonsocial environment. Overall, our results pave the way for future studies exploring the connection between contextual information and nonverbal behaviors for personality recognition from a psychological as well as computational point of view.

{\renewcommand{\arraystretch}{1.2}
\begin{table}
\caption{Experiments on the Salsa dataset for Personality Recognition. CV= Cross-validation, LOO= leave-one-out. The p-values less than 0.001 are summarized with three asterisks, p-values less than 0.01 are summarized with two asterisks, p-values less than 0.05 are summarized with one asterisk, and finally, no asterisks show not significant p-values.}
    \begin{center}
        \begin{tabular}{c|c|c|c|c|c|c}
            \hline
            Methods & \shortstack{Salsa\\Poster\\CV} & \shortstack{Salsa\\Party\\CV} &\shortstack{\\Avg.\\CV} & \shortstack{Salsa\\Poster\\LOO}  & \shortstack{Salsa\\Party\\LOO} & \shortstack{\\Avg.\\LOO} \\
            \hline
            \multicolumn{1}{c}{\textbf{Baseline}}\\
            \hline
            $LSTM_{cl}$ \cite{dotti2018behavior} & $65.7\%$& $67.85\%$ & $66.77\%$ & $65.8\%$  & $53.38\%$ & $59.59\%$\\
            EL-LMKL \cite{beyan2018investigation} & $75\%$ & $74.75\%$ & $74.87\%$ & $73.6\%$ & $48.9\%$ & $61.25\%$  \\
            Clips+MTLN \cite{ke2017new} & $78.7\%$ & $74.5\%$ & $76.6\%$ & $78.1\%$ & $58.86\%$ & $68.48\%$ \\
            \hline
            \multicolumn{1}{c}{\textbf{Proposed}}\\
            \hline
            $D_{pers}$ & $76\%$ & $71.3\%$ &  $73.65\%$ & $76.4\%$ &  $59.66\%$ & $68.03\%$ \\
            $D_{prox}+D_{group}$ & $73.6\%$ & $75.83\%$ & $74.71\%$ & $75\%$ & $59.13\%$ & $67.06\%$ \\
            $D_{pers}+D_{prox}$ & $78.2\%$ & $75.61\%$ & $76.9\%$ &$78\%$ & $62.4\%$ & $70.2\%$ \\
            $D_{pers}+D_{group}$ & $79.6\%$ & $74\%$&  $76.8\%$ & $79\%$ & $64.93\%$ & $71.96\%$\\
            $D_{pers}+D_{prox}+D_{group}$ & \textbf{79.8\%} & \textbf{77.25\%*} & \textbf{78.52\%*} & \textbf{79.2\%}  & \textbf{66.8\%***} & \textbf{73\%**} \\
            \hline
        \end{tabular}
    \end{center}
\label{table:salsa_types}
\end{table}}

%-------------------------------------------------------------------------
{\renewcommand{\arraystretch}{1.2}
\begin{table}
\caption{Experiments on the Nonsocial dataset for Personality Recognition. CV= Cross-validation, LOO= leave-one-out. P-values less than 0.05 are summarized with one asterisk.}
    \begin{center}
        \begin{tabular}{ccc}
            \hline
            Methods & \shortstack{Nonsocial\\CV} & \shortstack{Nonsocial\\LOO}\\
            \hline
            \multicolumn{2}{c}{\textbf{Baseline}}\\
            \hline
            $LSTM_{cl}$ \cite{dotti2018behavior} & $66.4\%$& $55.3\%$ \\
            Clips+MTLN \cite{ke2017new} & $70.9\%$ & $70.7\%$  \\
            \hline
            \multicolumn{2}{c}{\textbf{Proposed}}\\
            \hline
            $D_{pers}$ &  $69.4\%$& $69.2\%$\\
            $D_{pers}+D_{prox}$ & \textbf{72.4\%*} & \textbf{72.6\%*} \\
            \hline
        \end{tabular}
    \end{center}

\label{table:nonsocial_types}
\end{table}}

{\renewcommand{\arraystretch}{1.5}
\begin{table}
\caption{Personality Recognition using Traits scores, E= Extraversion, A= Agreableness, C= Conscientiousness, N= Neuroticism, OE= Open to Experience}
    \begin{center}
        \begin{tabular}{cccc}
        \hline
        \shortstack{Personality\\Traits} & \shortstack{Salsa\\Poster} &\shortstack{Salsa\\Party} & Nonsocial\\
        \hline
        E & $86.6\%$ & $80.51\%$ & $68.6\%$ \\
        \hline
        A & $86.2\%$ & $75.7\%$ & $77.2\%$ \\
        \hline
        C & $82.8\%$ & $75.25\%$ & $77\%$ \\
        \hline
        N & $82.5\%$ & $80.23\%$ & $74.8\%$ \\
        \hline
        OE & $91\%$ & $78.15\%$ & $77\%$ \\
        \hline
        \end{tabular}
    \end{center}

\label{table:traits}
\end{table}}
%-------------------------------------------------------------------------

\subsection{Personality traits}
\label{sec:pers_rec_traits}

The Big-Five personality Traits \cite{mccrae1992introduction} is the most popular model for personality recognition, hence, it is of great importance that the proposed framework is evaluated using the big five traits labels. To the best of the authors' knowledge, this is the first study presenting the recognition of personality traits on the two chosen datasets, providing new insights on personality patterns in the depicted scenarios.

In order to use the traits scores as classification labels, following the approach used by \cite{zen2010space}, the score of each trait is transformed into a binary value (HIGH or LOW), based on the median value computed from the population. Note that every  trait is evaluated independently from the scores of the other traits. As for the proposed architecture, we employ the descriptors that obtained the best results in the previous experiment. Hence, the $D_{pers}+D_{group}+D_{prox}$ framework is used for the Salsa dataset, and the $D_{pers}+D_{prox}$ framework is used for the Nonsocial dataset.

Results of the K-fold cross validation experiment are reported in Table \ref{table:traits}. For the Salsa dataset, like in the previous experiment, results are presented for the two parts of the dataset independently. In the poster session part, the highest accuracy is obtained for the $OE$ trait, followed by $E$ and $A$ traits. Considering the scenario in which the dataset was recorded (university environment), behavioral attributes related to these traits, such as curiosity or being talkative, can be easily matched with researchers' behaviors depicted in the video data. On the other hand, in the cocktail party part, the highest accuracy is obtained for the $E$, as well as $N$ traits. As in this scenario the participants are allowed to freely interact with each other, attributes like talkative and sociable belonging to the $E$ trait, are more evident than in the other scenario. For the nonsocial dataset, which contains problem solving tasks data for ADL applications, the highest accuracy is obtained for the $A$, $C$ and $OE$ traits. When it comes to engagement with the scene, searching for objects, deliberate scanning strategies or curiosity are attributes related to $C$ and $OE$ traits, which can be matched to behaviors observed in the nonsocial dataset. Moreover, given the nonsocial scenario, as expected, the $E$ trait is the hardest to predict.

%-------------------------------------------------------------------------

\begin{figure*}[!t]
\centering
\includegraphics[height=9cm,width=14cm]{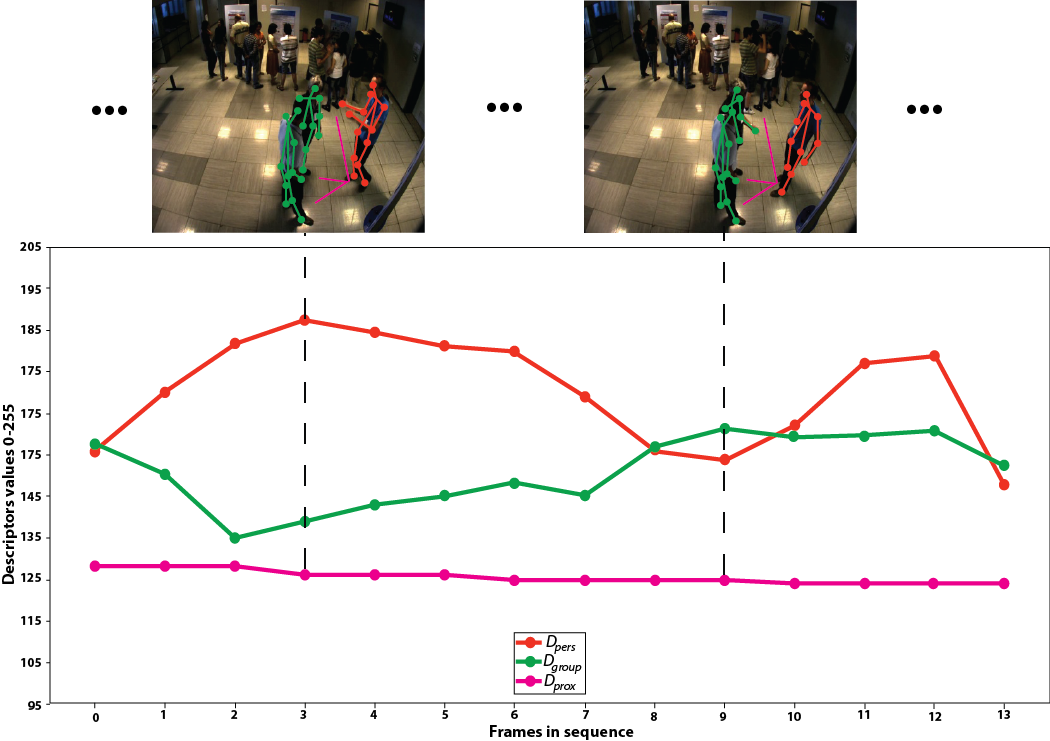}
\caption{Example of correctly predicted sequence, belonging to high $A$ trait. The graph contains the descriptors values (depicted in red, green and purple) on the y-axis, for each frame in the sequence on the x-axis. An alternation of high values is noticeable between the individual motion descriptor $D_{pers}$ and the social group motion $D_{group}$, indicating that the two parts are taking  turns in the conversation. Correspondingly, the image at frame $3$ shows high motion of the skeleton colored in red, and the image at frame $9$ shows high motion of the skeleton colored in green.}
\label{fig:good_example}
\end{figure*}

\section{Qualitative Results}
\label{sec:qual_res}

\begin{figure*}[!t]
\centering
\includegraphics[height=9cm,width=14cm]{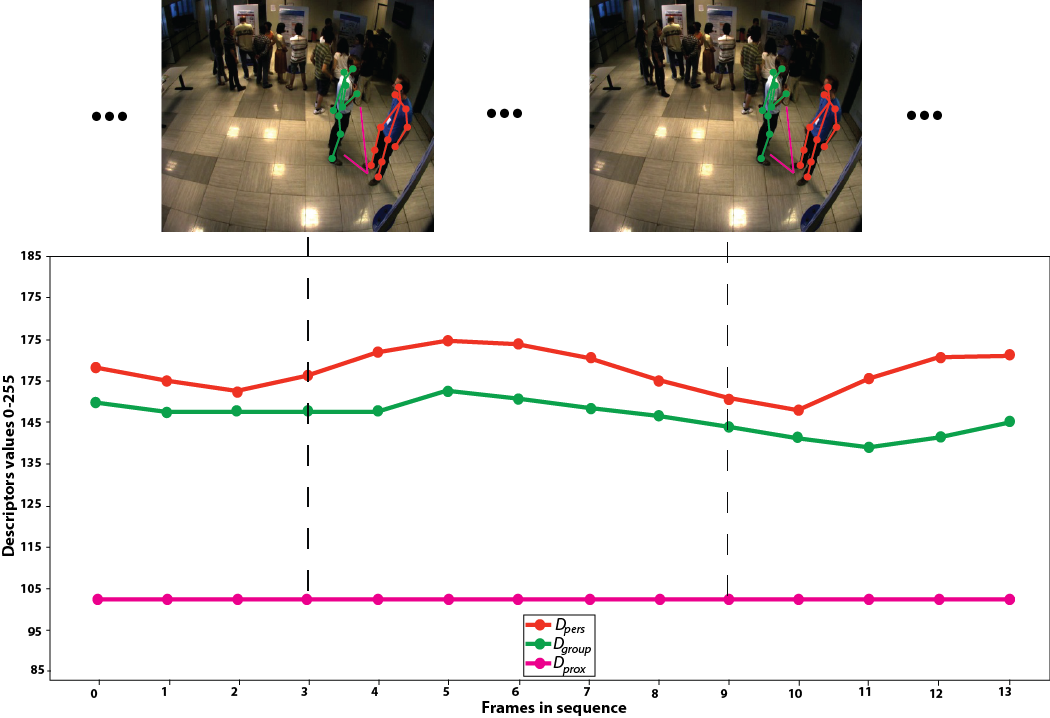}
\caption{Example of misclassified sequence belonging to high $A$ trait. The graph contains the descriptors values (depicted in red, green and purple) on the y-axis, for each frame in the sequence on the x-axis. As the interaction between the two individuals is limited, and the descriptors values are quite low, the system did not associate this behavioral pattern with high $A$ trait.}
\label{fig:bad_example}
\end{figure*}

As explained in section \ref{sec:experimets}, all the proposed descriptors contribute to the understanding of certain behavioral patterns, bearing to an improved personality recognition performance. In this section, aiming to further investigate the learning process, we disentangle the behavior of each descriptor at test time. Specifically, we select two frame sequences $t_1,t_2$ belonging to a subject with high Agreeableness trait. One sequence, ($t_1$) depicted in Figure \ref{fig:good_example} was classified correctly and the network showed high confidence that the sequence belonged to the right label, whereas the other sequence ($t_2$) shown in Figure \ref{fig:bad_example} was misclassified. The graphs display the three descriptor values before being fed to the CNN learning framework. On the y-axis, we show the individual motion values ($D_{pers}$ red line), social group motion ($D_{group}$ green line), and proxemics ($D_{prox}$ purple line) for each frame of the sequence, on the x-axis. 
In the correctly classified sample (Fig. \ref{fig:good_example}), we can notice an alternation of high values between the individual motion descriptor $D_{pers}$, and the social group motion descriptor $D_{group}$. This variation of high motion may denote that the two parts are conversing, each taking turns in a discussion. The displayed behavioral pattern, identified by our framework as belonging to the Agreeableness trait, is in line with previous literature studies, which correlate this trait with aspects like cooperation and empathy \cite{weisberg2011gender}.  

In the misclassified sample (Fig. \ref{fig:bad_example}), the interaction is more limited, and the descriptors values display low variation, while, the conversation occurs between only two individuals making the $D_{prox}$ values low as well. This pattern violates previous literature findings, which correlate high Agreeableness trait with large and highly interactive conversational groups \cite{alameda2015analyzing}. Hence, to further improve the results, our framework should be able to better model situations in which the motion is rather limited, or the behavior of the individuals is close to being ``static'', by including additional sources of information, such as facial or audio data.

\section{Discovered Personality Patterns}
\label{sec:patterns}

In this section, class activation maps are explored \cite{zhou2016learning} to exploit which discriminative personality patterns our proposed CNN framework learned in the social scenario data. Specifically, we use the  $D_{pers}+D_{group}$ descriptor on the Salsa dataset, to reveal the interaction between individual and group behaviors with different personality traits. The activation of the feature maps extracted by the VGG component, corresponds to spatial information of the original image \cite{ke2017new}. We aim to discover the importance (determined by the classifier weights in the softmax layer) of both individual dynamics as well as the social group dynamics for each personality trait, by applying the class activation maps on the Person-Context descriptors.

\begin{equation}
\label{eq:sptial_map}
    \begin{split}
        Act_p(x,y) = \sum_{k=1}^{K}{w_{k}^{c}f_k(x,y)}, k \in{1,\ldots,K},\\ K=D_{pers}+D_{group}
    \end{split}
\end{equation}
, where $w$ is the weights vector of the softmax layer, $f_k(x,y)$ is the activation of the $k$ unit in the last convolutional layer at spatial location $(x,y)$ for the personality class $c$, while the number of units $K$ is formed of two components, one corresponding to the Person information ($D_{pers}$) and one to the Context information ($D_{group}$).

\begin{figure*}[!t]
\centering
\includegraphics[scale=0.49]{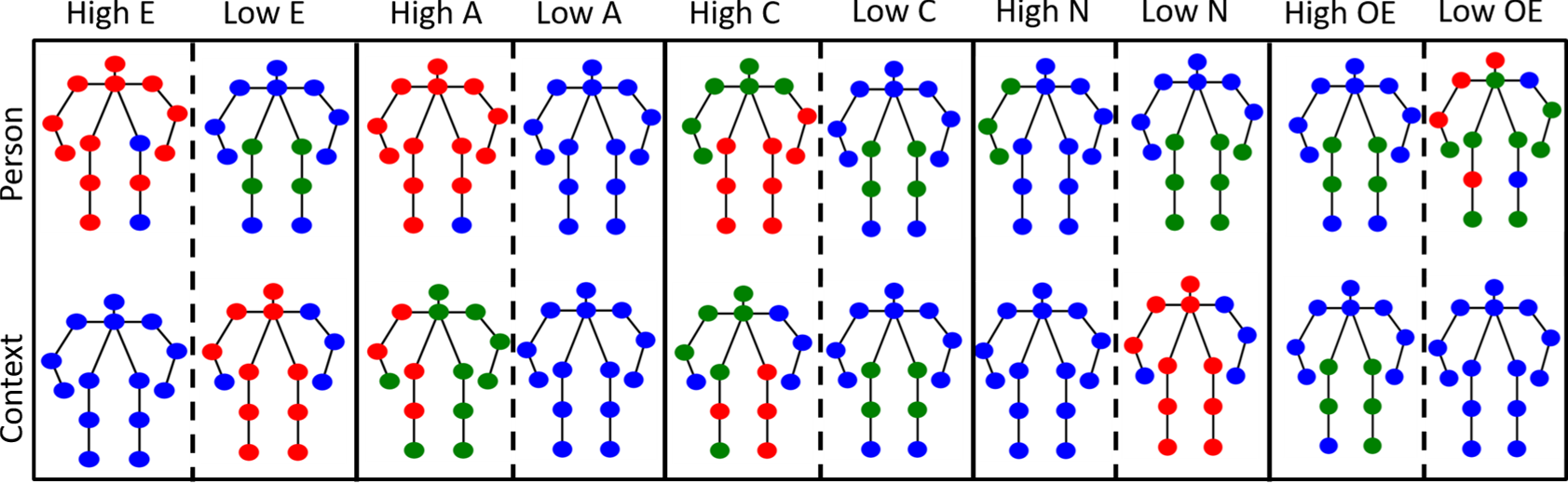} %height=6cm,width=15cm

\caption{Discovered personality patterns using Person-Context descriptors, given the \textbf{High} and \textbf{Low} scores of each personality trait: \textbf{E}xtroversion, \textbf{A}greeableness, \textbf{C}onscientiousness, \textbf{N}euroticism and \textbf{O}penness to \textbf{E}xperience. Low CNN weights activation in blue, medium CNN weights activation in green, and high CNN weights activation in red.}
\label{fig:personality_patterns}
\end{figure*}

We are interested in visualizing meaningful personality patterns for each personality trait. Specifically, we pose the following questions: 1) \textit{Is the relation between Person-Context dynamics reflected in the traits?} and 2) \textit{Do the dynamics correspond to the trait attributes defined by psychologists?} The discovered personality patterns for each trait are displayed in Fig. \ref{fig:personality_patterns}. In particular, we visualize the learned patterns from both Person and Context CNN networks. We quantize the activation of each joint in three groups: low activation (first quartile) depicted in blue color, medium activation (second quartile) depicted in green color, and finally, high activation (third quartile) is depicted in red color.
When an individual has a high $E$ trait (e.g. talkative, outgoing), the CNN weights show high activation on the features coming from the Person stream, and low activation in the features coming from the Context stream. As a consequence, if an individual has a low E trait, features coming from the Context stream are more important. Thus, to answer the first question, the two stream dynamics are learned by the CNN architecture and are mapped to personality behavioral patterns. In order to answer the second question, we highlight that, since individuals with a high $E$ trait are described as talkative and outgoing, when it comes to social interaction, they tend to be the \textit{center of attention}, forcing the rest of the social group to be more passive \cite{lu2009size}. On the other hand, CNN weights for individuals with high $A$ and $C$ traits, show high activation on both Person-Context streams, demonstrating that these personality types like to be engaged with the social group, without trying to dominate the situation \cite{mccarty2005personality}. Medium/high CNN activation on both Person-Context streams is found also for individuals with low $N$ and $OE$ traits, showing that having low scores in attributes, such as being tense or being imaginative, makes individuals more sociable.

\section{Conclusion}

In this paper, we have presented a novel CNN-based framework for personality recognition. Our model analyzes the scene at multiple levels of granularity: firstly, we encode spatio-temporal descriptors for each individual in the scene, secondly, we extract  spatio-temporal descriptors from social groups, and thirdly, we encode the global proxemics of every individual in the scene. Additionally, we demonstrate that our proxemics features can be applied also in a nonsocial scenario, encoding scene interaction information. Experiments on two personality recognition datasets demonstrate the effectiveness of our approach, showing that modeling together Person-Context information significantly improves the state-of-the-art personality recognition results. Furthermore, we presented CNN class activation maps for each personality trait, providing novel insights on nonverbal behavioral patterns linked with personality attributes defined by theories from behavioral psychology. We believe these findings are of great importance for future interdisciplinary behavioral studies, aiming to combine data-driven approaches with psychological studies.

\begin{acks}
This work has been funded by the European Union' Horizon 2020 Research and Innovation Programme under Grant Agreement $N^{\circ}$ 690090 (ICT4Life project).
\end{acks}
  
% \section{Acknowledgments}

% Identification of funding sources and other support, and thanks to individuals and groups that assisted in the research and the preparation of the work should be included in an acknowledgment section, which is placed just before the reference section in your document. 

% This section has a special environment:
% \begin{verbatim}
%   \begin{acks}
%   ...
%   \end{acks}
% \end{verbatim}
% so that the information contained therein can be more easily collected during the article metadata extraction phase, and to ensure consistency in the spelling of the section heading. 

% Authors should not prepare this section as a numbered or unnumbered {\verb|\section|}; please use the ``{\verb|acks|}'' environment.

% The next two lines define the bibliography style to be used, and the bibliography file.
\bibliographystyle{ACM-Reference-Format}
\bibliography{person-context}

\end{document}